\documentclass{article}

\usepackage{arxiv}

\usepackage[utf8]{inputenc} 
\usepackage[T1]{fontenc}    
\usepackage{hyperref}       
\usepackage{url}            
\usepackage{booktabs}       
\usepackage{amsfonts}       
\usepackage{nicefrac}       
\usepackage{microtype}      
\usepackage{lipsum}		
\usepackage{graphicx}
\usepackage{doi}

\usepackage[ruled,vlined]{algorithm2e}
\usepackage{multirow}
\usepackage{subcaption}
\usepackage{pifont}
\usepackage{amsmath}
\usepackage{amssymb}
\newcommand{\cmark}{\ding{51}}%
\newcommand{\xmark}{\ding{55}}%

\title{Robust Object Detection in Remote Sensing Imagery with Noisy and Sparse Geo-Annotations (Full Version)}

\author{Maximilian Bernhard\\
    LMU Munich \\
	\texttt{bernhard@dbs.ifi.lmu.de} \\
	\And
    Matthias Schubert \\
    LMU Munich \\
    \texttt{schubert@dbs.ifi.lmu.de}
}

\renewcommand{\headeright}{}

\hypersetup{
pdftitle={Robust Object Detection},
pdfsubject={},
pdfauthor={Bernhard and Schubert},
pdfkeywords={computer vision, remote sensing, object detection, noisy annotations, missing annotations, localization errors}
}

\begin{document}
\maketitle

\begin{figure}[h!] 
    \centering
  \includegraphics[width=0.49\textwidth]{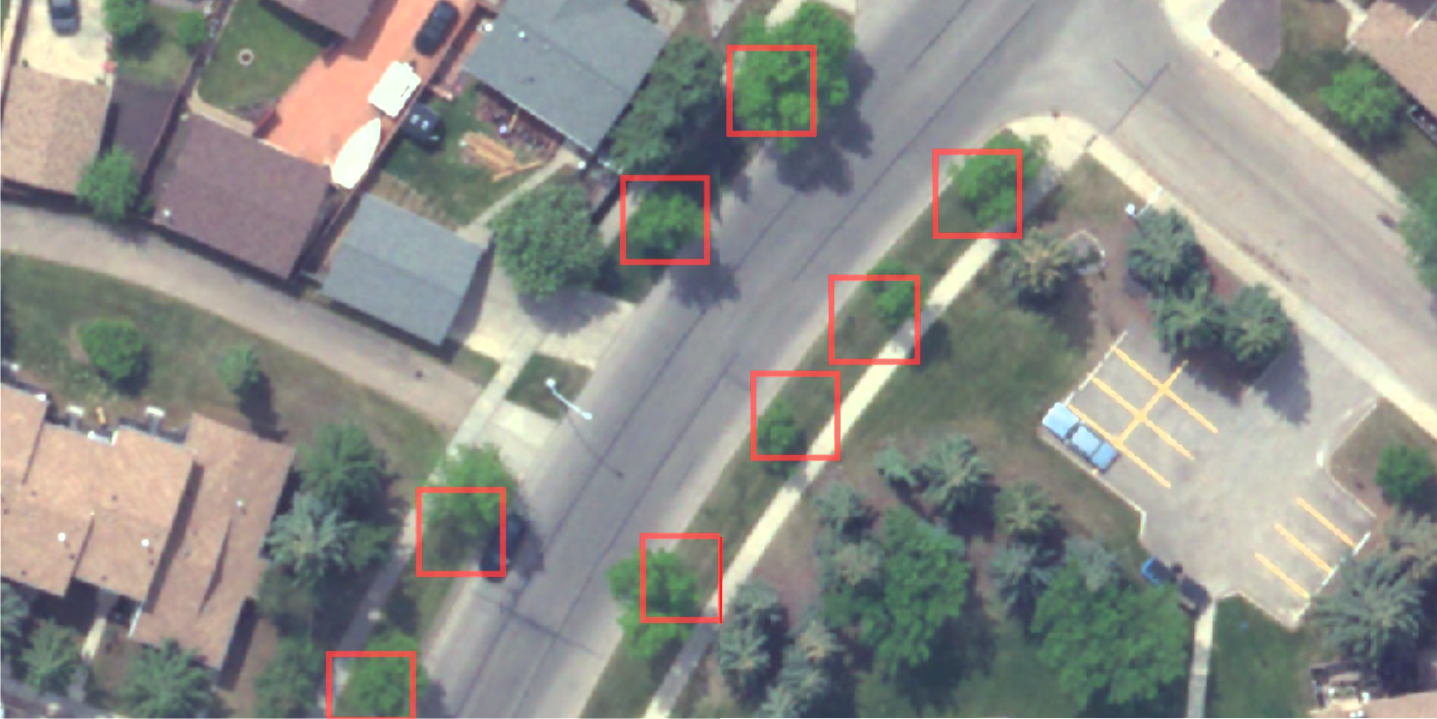}
  \includegraphics[width=0.49\textwidth]{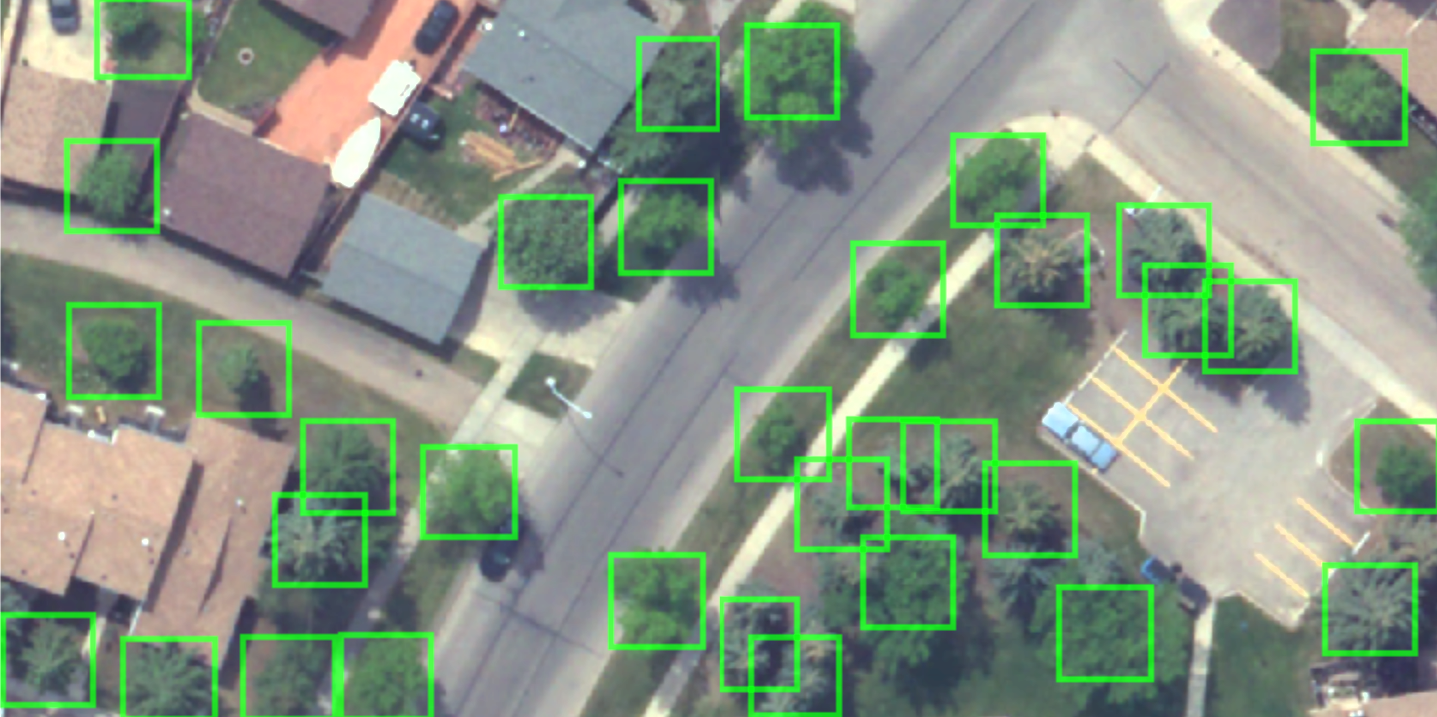}
  \caption{Available noisy and incomplete annotations (left) vs. final predictions with our method (right) for a real-world dataset covering urban trees. \emph{Best viewed in color.}}
  \label{fig:teaser}
\end{figure}

\begin{abstract}
Recently, the availability of remote sensing imagery from aerial vehicles and satellites constantly improved. For an automated interpretation of such data, deep-learning-based object detectors achieve state-of-the-art performance. However, established object detectors require complete, precise, and correct bounding box annotations for training. In order to create the necessary training annotations for object detectors, imagery can be georeferenced and combined with data from other sources, such as points of interest localized by GPS sensors. Unfortunately, this combination often leads to poor object localization and missing annotations. Therefore, training object detectors with such data often results in insufficient detection performance.
In this paper, we present a novel approach for training object detectors with extremely noisy and incomplete annotations.
Our method is based on a teacher-student learning framework and a correction module accounting for imprecise and missing annotations. Thus, our method is easy to use and can be combined with arbitrary object detectors. 
We demonstrate that our approach improves standard detectors by 37.1\% $AP_{50}$ on a noisy real-world remote-sensing dataset. Furthermore, our method achieves great performance gains on two datasets with synthetic noise. Code is available at \url{https://github.com/mxbh/robust_object_detection}.
\end{abstract}


\section{Introduction}
In the last decade, the quality and quantity of available aerial and satellite imagery has considerably increased due to an easier and cheaper acquisition. At the same time, methods for object detection in images have been tremendously improved with the rapid advancements in the field of deep learning. As a consequence of these developments, deep-learning-based object detectors have established themselves as an essential and highly valuable tool for the automated analysis of remote sensing imagery. For instance, object detection was successfully used for monitoring vegetation~\cite{rs-seedling-detection,rs-banana}, detecting natural hazards~\cite{rs-wildfire-detection,rs-fire-detection}, urban planning~\cite{acm-obj-det-traffic,acm-ua-obj-det}, and many more applications~\cite{rs-obj-det-survey2,rs-obj-det-survey3}.
Unfortunately, object detectors require a large number of annotated training samples. Furthermore, the quality of annotations is crucial for the final performance~\cite{ca-bbc,nar,co-mining}. Therefore, annotations are usually created manually by human annotators who draw bounding boxes around objects of interest and assign class labels. As this process is time-consuming and might -- depending on the use case -- require expert knowledge, the availability of sufficient annotations often becomes an obstacle in practice.

Thus, the optimal usage of already existing annotations is key toward reducing labeling efforts and improving the applicability of object detectors. In geospatial applications, georeferenced imagery can be combined with geolocalized annotations from a different source. For instance, GPS coordinates resulting from a ground survey may be later mapped onto drone imagery of the area. However, this procedure is usually faulty, leading to two common types of annotation errors: First, aligning GPS annotations with georeferenced imagery often results in certain misalignments and displacements between images and annotations due to inevitable factors like GPS errors, uneven terrain, or changing camera altitudes and angles during image acquisition~\cite{georect-hughes,georect-rio,georect-gomez,rs-image-alignment}. 
Also, if GPS coordinates only provide points instead of rectangular bounding boxes, the boxes have to be created artificially by assuming a fixed, square shape for every object. The reason for this is that object detectors require bounding boxes for supervision and it might clearly lead to badly fitting boxes.
Therefore, models trained with such imprecise supervision may struggle to learn to localize the objects of interest accurately. Depending on the severity of localization errors, this might also harm the detector's ability to detect the objects at all.
Second, the completeness of GPS annotations is often an issue as not all objects of interest might have been registered. For example, in urban areas, objects on private ground cannot be registered in field surveys. Also, temporal differences between the acquisition of images and annotations can lead to such label errors. This false negative supervision can cause detectors to wrongly predict present objects as background. False positive supervision caused by superfluous annotations may also occur, but we do not address this phenomenon in this work as we found it to be negligible (see Appendix~\ref{sec:app-superfluous}).

Both of these effects, i.e. imprecise and missing annotations, can be clearly observed in a public inventory covering urban trees in the city of Edmonton, Canada (see left-hand side of Figure~\ref{fig:teaser}). Combining these GPS readings with aerial images results in an extremely noisy and thus challenging object detection dataset, subsequently referred to as \emph{Edmonton Trees}. While standard methods suffer severely under the low label quality in this dataset, our approach produces strong results (see right-hand side of Figure~\ref{fig:teaser} and Section~\ref{sec:main-results}).

So far, comparatively few efforts have been put into solving the problems caused by corrupted annotations in the context of object detection. On the one hand, there is a line of research for dealing with noisy bounding boxes and incorrect class labels~\cite{ca-bbc,nar,co-correction,sd-locnet,note-rcnn}. On the other hand, there is some work on \emph{Sparsely Annotated Object Detection} (SAOD)~\cite{co-mining,brl,soft-sampling,saod-region-based}, i.e. object detection with incomplete annotations. In comparison to \emph{Semi-supervised Object Detection} (SSOD), there is no fully annotated training subset in SAOD and all of the training images are potentially incompletely labeled. Thus, every negative supervision might be incorrect in SAOD, posing an obstacle for training. 
However, none of these methods deal with the combination of localization noise and annotation sparsity. In~\cite{noisy-segmentation,map-repair}, methods for semantic segmentation with inaccurate and incomplete annotations are proposed, while there is no comparable work addressing object detection under such adverse conditions.

To fill this gap, we present a unified framework for training object detectors with extremely noisy and sparse, i.e. incomplete, supervision. At the core of our approach is a correction module that takes model predictions and noisy annotations as input and produces refined targets for stable training. Thus, it operates in an unsupervised way without any need for clean annotations. The module consists of two independent submodules, one for the correction of noisy bounding boxes and one for the correction of missing annotations. Our correction module is isolated from the detector architecture and the loss function, making it simple and easy to extend. Similar to the approach for semi-supervised object detection proposed in~\cite{unbiased-teacher}, we employ a teacher-student learning framework, in which a student is supervised by targets that have been corrected by a teacher network. In doing so, we achieve a robust training scheme that is able to cope with extremely low annotation quality. We conduct experiments on the aforementioned Edmonton Trees dataset, exhibiting grave real-world annotation noise. 
Additionally, we analyze the effects of different severities of synthetic noise on the commonly used object detection datasets NWPU VHR-10~\cite{nwpu} and Pascal VOC~\cite{pascal-voc}.
Particularly, Pascal VOC, a highly popular dataset in the computer vision community, is used to compare our box correction method with other approaches for this task.

To summarize, our contribution is threefold: 
\begin{itemize}
  \item We propose a novel correction module, accounting for imprecise object localization and missing annotations during training.
  
  \item We integrate our correction module into a teacher-student training framework. In doing so, we can greatly improve the performance of standard detectors on the real-world Edmonton Trees dataset as well as on NWPU VHR-10 and Pascal VOC with synthetic noise. To the best of our knowledge, we are the first to tackle the task of object detection when solely extremely imprecise and sparse annotations are available.
  \item With our framework and our box correction module, we beat several state-of-the-art approaches on Pascal VOC with synthetic bounding box noise.
\end{itemize}

\section{Related Work}

\paragraph{Object Detection for Geospatial Applications and Remote Sensing}
There are a plethora of works in the geospatial domain, where object detection is used for image analysis. The application areas cover a broad range comprising, for example, urban planning, transportation, vegetation and wildlife monitoring, agriculture, natural hazard detection, and GIS updating~\cite{rs-obj-det-survey3,rs-obj-det-survey2,rs-obj-det-survey1,acm-ua-obj-det}.
In general, remote sensing imagery and the contained objects exhibit certain characteristics that are different from natural scene images. Therefore, both a line of research specifically dealing with remote sensing and geospatial object detection ~\cite{nwpu,acm-ship-detection, acm-ua-obj-det,rs-obj-det3,rs-obj-det4} as well as a variety of remote sensing datasets for benchmarking object detectors~\cite{nwpu,rs-data-cowc,rs-obj-det-survey2,rs-obj-det-survey3} arose. In order to get clean training supervision, most of these datasets were annotated manually~\cite{rs-obj-det-survey2}.

\paragraph{Semi-supervised and Weakly Supervised Object Detection}
Methods for semi-supervised object detection (SSOD)~\cite{stac,unbiased-teacher,soft-teacher} and weakly supervised object detection (WSOD)~\cite{wsod1,wsod2,wsod3,wsod4} aim at training object detectors without a complete training set and complete annotations, thereby reducing labeling requirements. In semi-supervised object detection, a set of samples with clean annotations is supplemented with a larger set of completely unlabeled samples. In contrast, no clean annotations at all are available in our setting, making semi-supervised detection methods unsuitable for our problem.
In weakly supervised learning, labels of lower quality than the expected predictions are used in general. Therefore, our task can be seen as a variant of weakly supervised object detection. However, most works on weakly supervised object detection solely employ image-level class labels. Consequently, the performance of such methods is generally far behind fully supervised methods, whereas our method is able to largely close the gap to fully supervised models in settings with moderate noise.

\paragraph{Classification with Noisy Labels}
In~\cite{rethinking-generalization}, it has been shown that deep neural networks can easily fit random labels. Thus, random and corrupted labels can have a detrimental impact on learning and generalization. Consequently, different methods such as~\cite{co-teaching,mentornet,dividemix,nested-co-teaching,robust-loss-functions} that aim to mitigate the effects of noisy supervision have been developed. 
Although these methods are rather general as they were proposed for the task of classification, their applicability to more complex tasks such as object detection is limited. In object detection, label noise cannot only corrupt class labels of objects but also bounding boxes. Hence, a dedicated line of research evolved for object detection with corrupted labels. Our work follows this direction.

\paragraph{Object Detection with Noisy Annotations}
The approach proposed in~\cite{ca-bbc} aims for robust training of the Faster R-CNN object detector~\cite{faster-rcnn} under both class label noise and bounding box noise. 
Similarly and assuming the same setting, the authors of~\cite{nar} also propose a method to correct label noise and box noise.
Furthermore,~\cite{nlar,alpha-iou,co-correction} present approaches for dealing solely with inaccurate bounding boxes. 
More precisely,~\cite{co-correction} assumes a very specific setting where only point annotations are available. We extend this method by enabling end-to-end training, individual box coordinate correction, and adding a mechanism to deal with incomplete annotations.
In addition, there are some works that were originally proposed for different tasks, such as weakly-supervised~\cite{sd-locnet} (image-level supervision) and semi-supervised~\cite{note-rcnn} object detection, but present useful techniques for noise-robust object detection as well.

\paragraph{Sparsely Annotated Object Detection}
The problem of Sparsely Annotated Object Detection (SAOD) can be seen as a special case of object detection with annotation noise, where some objects are not annotated or incorrectly labeled as background.
Existing approaches for SAOD follow two basic paradigms. On the one hand, one can mitigate the effect of false negatives during training by downweighting background losses that are likely to be caused by missing annotations~\cite{soft-sampling}. On the other hand, one can treat hard negatives as positives during loss computation if the model is confident about its foreground prediction~\cite{brl,co-mining}. Recently,~\cite{saod-region-based} proposed a semi-supervised approach for SAOD. In this work, we combine the pseudo-labeling scheme of~\cite{co-mining} with our box correction method in order to deal with sparse and inaccurate annotations simultaneously.

\paragraph{Object Detection with Sparse Annotations and Noisy Bounding Boxes}
Although being relevant in practice, the problem of object detection with extremely noisy and sparse annotations was hardly addressed in the past.
In~\cite{registree}, the authors report annotation noise, resulting in inaccurate localizations and missing annotations. However, they do not specifically tackle this issue, but rather focus on the integration of street view images into their learning system.
Recently, a method including mechanisms for these annotation error types was proposed in~\cite{mixtraining}, but it was only developed and evaluated in the standard setting on the COCO dataset~\cite{coco}, which is generally considered clean.
In contrast, we investigate extremely noisy settings, where standard methods do not suffice to cope with the annotation noise and fail.
Moreover, \cite{map-repair,noisy-segmentation} present approaches for dealing with noisy and partial labels for semantic segmentation in remote sensing applications, whereas we are interested in the task of object detection.

\paragraph{Pseudo-Labels}
According to~\cite{kaggle-pseudo-labels}, the usage of pseudo-labels for model training has the effect of entropy regularization and leads to low-density separation of classes.
Hence, pseudo-labels have been successfully employed in many settings where full or accurate supervision is not available, e.g. in semi-supervised object detection~\cite{unbiased-teacher,soft-teacher,stac}, unsupervised-domain adaptation~\cite{category-dictionary,uda-pseudo-labels} or weakly-supervised object detection~\cite{pcl,ts2c}. However, pseudo-labels come with the risk of self-confirmation bias and accumulating errors. To mitigate this, teacher-student~\cite{unbiased-teacher,soft-teacher} or co-training methods~\cite{co-mining,co-teaching} are often utilized. As corrected targets in our setup can also be seen as a form of pseudo-labels, we adopt this technique for our problem setting.

\section{Problem Setting}
Given a dataset consisting of images $I \in \mathbb{R}^{3 \times H \times W}$, our goal is to train an object detector $\Phi$ that is able to predict the true class label $l=1...L$ and a minimal bounding box $b=(x_1,y_1,x_2,y_2)$ for every object of interest in an image. Furthermore, the detector produces a probability score $s \in [0,1]$ indicating its confidence for every predicted instance. Thus, the detector output is a set of triplets of bounding boxes and class labels and confidence scores, i.e. $\Phi(I) = \big\{(\hat{b}_p, \hat{l}_p, \hat{s}_p)\big\}_{p =1...N_{preds}}$.
Instead of the true boxes and labels $\big\{(b_t, l_t)\big\}_{t =1...N_{true}}$, only a subset consisting of imprecise boxes $\big\{(\tilde{b}_t, \tilde{l}_t )\big\}_{t =1...N_{targets}}$ is available for supervision during training. That is, $N_{targets} \neq N_{true}$ as some target boxes are missing and, additionally, the coordinates of the available boxes do not describe the true extents of objects. In practice, superfluous annotations for image regions with no objects may be observed as well. However, we do not consider this issue as we argue that this occurs far less frequently than missing annotations and, furthermore, we found the effects of such erroneous annotations to be insignificant (see Appendix~\ref{sec:app-superfluous}).
If no bounding boxes but only single points of interest are available for $\tilde{b}_t$, we assume square-shaped boxes of a fixed size to conform with the standard input format for state-of-the-art object detectors.

\section{Method}

We propose a training framework, which employs a correction module that takes unrefined targets as well as predictions of a detector $\Phi$ for a given image as an input. Guided by the predictions, the correction module transforms potentially noisy targets into a new and more reliable set of targets. 
As we address two types of annotation errors, missing annotations and imprecise boxes, our solution comprises two submodules, each tackling one type of error. 

In the following, we explain the general training framework used in our experiments. Afterward, we describe the aforementioned correction submodules.

\subsection{Training Framework}
\label{sec:training}

\begin{figure}[t!]
    \centering
    \includegraphics[width=0.7\textwidth]{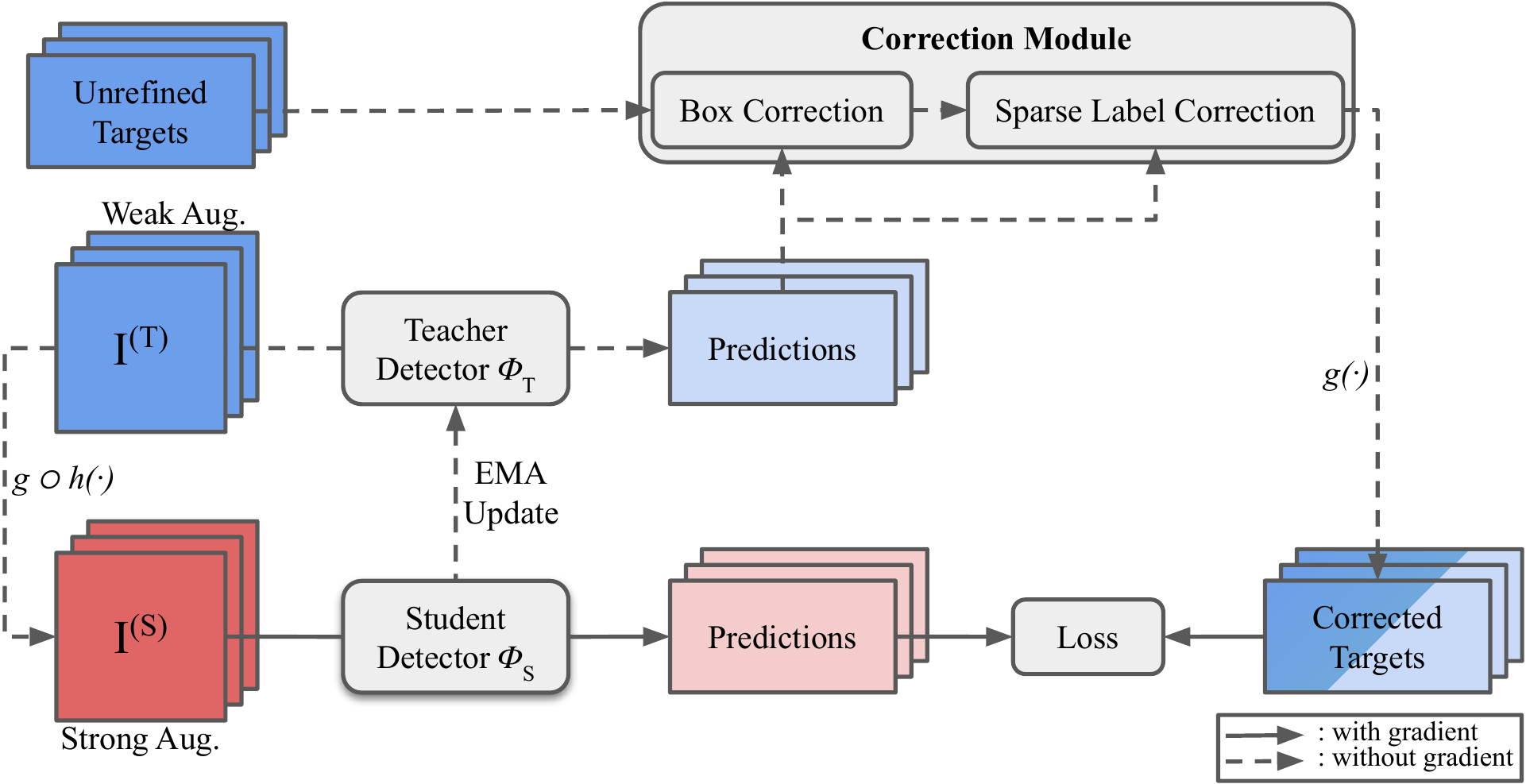}
    \caption{Our teacher-student training framework. \emph{Best viewed in color.}}
    \label{fig:framework}
\end{figure}

We perform a form of teacher-student training. This is a common practice that reduces self-confirmation bias when models (partly) supervise themselves. In our case, self-supervision is employed to mitigate the effect of annotation noise. 
However, when self-supervision is erroneous, plain training may lead to an amplification and accumulation of these errors, ultimately resulting in unstable training and bad generalization.
An asymmetric teacher-student architecture such as ours is an effective means to counter these problems.
In principle, our teacher-student framework is similar to the one proposed in~\cite{unbiased-teacher}.
Nonetheless, we additionally integrate our correction module to the framework. This is necessary as we solve a different task than~\cite{unbiased-teacher}, which was proposed for semi-supervised object detection. 
An overview of our training pipeline can be seen in Figure~\ref{fig:framework}.

For every image $I$, we start by creating a weakly augmented version $I_T$ and a strongly augmented version $I_S$ of it. Thereby, we assume $I_S = g \circ h (I_T)$, where $h(\cdot)$ denotes a photometric transformation that can consist of operations like color jitter, blurring, or brightness modification. In contrast, $g(\cdot)$ denotes a geometric transformation such as rotation or flipping. As weak augmentations, i.e. for creating $I_T$, we solely employ random resizing and flipping.

We feed $I_T$ and $I_S$ into a teacher detector $\Phi_T$ and student detector $\Phi_S$, respectively. Thereafter, the teacher predictions $\Phi_T(I_T) = \big\{(\hat{b}_p, \hat{l}_p, \hat{s}_p)\big\}_{p =1...N_{preds}}$ as well as the unrefined, noisy annotations $\big\{(\tilde{b}_t, \tilde{l}_t )\big\}_{t =1...N_{targets}}$ are used to create corrected targets $C = \big\{(c_t, l_t)\big\}_{t =1...N_{corr}}$ by applying our proposed correction module. 
Based on these refined targets, the loss for the student detector is computed, i.e.
$$
\mathcal{L}_{student} = \mathcal{L}\left( \Phi_S (I_S), g(C)\right).
$$
Here, $\mathcal{L}$ denotes the loss function defined by the detector architecture. Also, note that we have to geometrically align the corrected targets $C$ with the outputs of $\Phi_S$ as they were produced based on different image versions.
While $\Phi_S$ is continuously updated with $\mathcal{L}_{student}$ and a suitable variant of SGD, the teacher model $\Phi_T$ is an exponential moving average (EMA) of the student. That is, after every iteration, we update every weight parameter $\theta_T$ of the teacher as a convex combination of itself and its corresponding student parameter $\theta_S$, i.e. 
$$
\theta_T \leftarrow \alpha \theta_T + (1 - \alpha) \theta_S,
$$
where $\alpha$ is a keep rate in $[0,1]$. 
Before we start with our teacher-student training, a detector is trained in a standard supervised manner on the unrefined data to provide a good initialization for the teacher and the student model.
After training, the teacher $\Phi_T$ is used for inference.

\begin{algorithm}[t!]
\SetAlgoLined
\SetKwInOut{Input}{Input}
\SetKwInOut{Output}{Output}
\SetKwInOut{Params}{Params}

\Input{$\bullet$ Unrefined target boxes and class labels $\big\{(\tilde{b}_t, \tilde{l}_t)\big\}_{t =1...N_{targets}}$, where $\tilde{b}_t=\left(\tilde{x}_{1_t}, \tilde{y}_{1_t}, \tilde{x}_{2_t}, \tilde{y}_{2_t}\right)$ and \\ \quad $\tilde{l}_t\in \{1...L\}$\\
$\bullet$ Predicted boxes, class labels, and raw logit scores $\big\{(\hat{b}_p, \hat{l}_p, \hat{s}_p)\big\}_{p =1...N_{preds}}$ before postprocessing with \\ \quad NMS, where $\hat{b}_p=\left(\hat{x}_{1_p}, \hat{y}_{1_p}, \hat{x}_{2_p}, \hat{y}_{2_p}\right)$, $\hat{l}_p\in \{1...L\}$, and $\hat{s}_p\in \mathbb{R}$}
\Params{Box distance measure $\delta$, distance limit $d$, softmax temperature $\gamma$}
\Output{Corrected targets $C_B = \big\{(c_t, l_t)\big\}_{t =1...N_{targets}}$}
\hrulefill

$C_B = \emptyset$

\For{Class label $l = 1...L$}{

Determine targets and predictions belonging to class $l$ (as indices): \hfill (i)\\
$\mathcal{T}^{(l)} = \{t = 1...N_{targets}: \tilde{l}_t=l\}$ \\
$\mathcal{P}^{(l)} = \{p = 1...N_{preds}: \hat{l}_p=l\}$\\
Initialize corrected boxes: $c_t = \tilde{b}_t$ for $t \in \mathcal{T}^{(l)}$\hfill(ii)

    \While{not converged}{
        \For{$t \in \mathcal{T}^{(l)}$}{
            $\mathcal{J}_{c_t} = \{ p \in \mathcal{P}^{(l)}:
            t = \underset{t' \in \mathcal{T}^{(l)}}{\mathrm{argmin}} ~\delta(c_{t'},\hat{b}_p)\land\: \delta(\tilde{b}_{t}, \hat{b}_p) \le d \}$\hfill(iii)
        }
        \For{$t \in \mathcal{T}^{(l)}$}{
            \If{$\mathcal{J}_{c_t} \neq \emptyset$}{
            $\left(w_p\right)_{p \in \mathcal{J}_{c_t}} = \mathrm{softmax}\left((\hat{s}_p / \gamma)_{p \in \mathcal{J}_{c_t}}\right)$
            
            $c_t = \left(\sum_{p \in \mathcal{J}_{c_t}} w_p \cdot \hat{\square}_{p}\right)_{\square=x_1,y_1,x_2,y_2}$\hfill(iv)
            }
    
        }
     }
     Add $\{(c_t,l): t \in \mathcal{T}^{(l)}\}$ to $C_B$
 }
 \Return{$C_B$}
 \caption{Box Correction}
 \label{algo:box-correction}
\end{algorithm}

\subsection{Box Correction}
\label{sec:box-correction}
To correct potentially noisy boxes, we apply Algorithm~\ref{algo:box-correction} using the teacher predictions. The main idea is to determine predicted boxes that are sufficiently overlapping with an unrefined target box and then replace it with a weighted average of the overlapping predicted boxes. 

More precisely, we start by selecting the noisy target boxes $\mathcal{T}^{(l)}$ and teacher predictions $\mathcal{P}^{(l)}$ that have a certain class label~$l$~(i). Next, we apply an iterative, local averaging scheme that is similar to k-Means where the centroids correspond to the corrected boxes. We initialize the corrected boxes $c_t$ as copies of the unrefined target boxes $\tilde{b}_t$ for $t \in \mathcal{T}^{(l)}$(ii). In each iteration, we assign each predicted box $\hat{b}_p$ for $p \in \mathcal{P}^{(l)}$ to a corrected box $c_t$ if it meets two conditions specified in 
\begin{equation}
\mathcal{J}_{c_t} = \Big\{ p \in \mathcal{P}^{(l)}: t = \underset{t' \in \mathcal{T}^{(l)}}{\mathrm{argmin}}~\delta(c_{t'},\hat{b}_p) \land \delta(\tilde{b}_{t}, \hat{b}_p) \le d \Big\}  \tag{iii}
\end{equation}
First, the corrected box $c_t$ has to be closer to the predicted box $\hat{b}_p$ than any other corrected box $c_{t'}$. As a suitable distance measure for rectangular bounding boxes, we use $\delta(b_1, b_2) = 1 - IoU(b_1, b_2) \in [0,1]$, where $IoU(\cdot,\cdot)$ denotes the intersection over union. Let us note that if this choice does not suit the characteristics of the dataset, another measure such as $GIoU$~\cite{giou} can be used.
The second condition for the assignment of a predicted box $\hat{b}_p$ to a corrected box $c_t$ is that it is sufficiently close to the original, noisy target box $\tilde{b}_t$, i.e. their distance $\delta(\tilde{b}_t,\hat{b}_p)$ may not exceed a threshold $d \in \mathbb{R}^+$. This design choice ensures that very distant predicted boxes (possibly false positives) do not impact the corrected boxes.

\begin{figure}[t!]
    \centering
    \includegraphics[width=0.5\columnwidth]{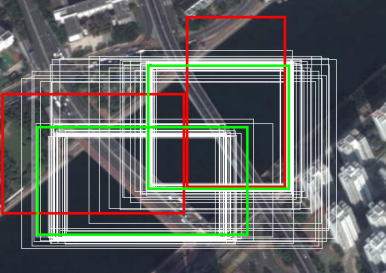}
    \caption{Illustration of the proposed box correction method. Initially noisy boxes (red) are iteratively updated as weighted averages of matching predicted boxes (white) to finally obtain corrected boxes (green). \emph{Best viewed in color.}}
    \label{fig:box_correction}
\end{figure}

After the assignment of predicted boxes, the corrected boxes $c_t$ are updated as the weighted average of their assigned predicted boxes, i.e. 
\begin{equation}
c_t = \left(\sum_{p \in \mathcal{J}_{c_t}} w_p \cdot \hat{\square}_{p}\right)_{\square=x_1,y_1,x_2,y_2},
\tag{iv} 
\end{equation}
where
$$
\left(w_p\right)_{p \in \mathcal{J}_{c_t}} = \mathrm{softmax}\left((\hat{s}_p / \gamma)_{p \in \mathcal{J}_{c_t}}\right).
$$
The average is computed for each of the four box coordinates $\square=x_1,y_1,x_2,y_2$ separately. To obtain the weights, we softmax the scores of the assigned boxes (note that, in contrast to the rest of the paper, we use raw logit scores instead of probability scores in Algorithm~\ref{algo:box-correction}). Thus, predicted boxes with high confidence receive larger weights. Moreover, we use the softmax temperature $\gamma$ as an additional hyperparameter that further allows controlling the impact of low-scoring boxes. We iterate assignments and updates until the corrected boxes do not change anymore and output the final corrected boxes denoted as $C_B = \big\{(c_t, \tilde{l}_t )\big\}_{t =1...N_{targets}}$. An illustration of this method can be seen in Figure~\ref{fig:box_correction}.

The advantage of this iterative scheme over a simple local averaging of boxes comes into effect when predicted boxes highly overlap with two or more noisy target boxes. In this case, the unique assignments and the iterative updates ensure that the corrected boxes correspond to a single target box and not to a blend of multiple boxes belonging to different objects. 
In cases where the given target boxes are well separated or only one target box is given, our method converges after a single update and degenerates to a local averaging scheme.

Altogether, this algorithm is similar to the correction scheme proposed in~\cite{co-correction}. However, there are some differences. In~\cite{co-correction}, the input to the algorithm is a set of points representing object locations instead of bounding boxes. Also, an activation map containing a score for every class and every pixel is needed, whereas our algorithm operates on bounding boxes and their confidence scores. Therefore, we can apply our algorithm to outputs of object detectors and integrate it into an end-to-end trainable framework. Furthermore, as we deal with bounding boxes, our method is not restricted to object center locations, but it is also able to correct bounding box extents.

\subsection{Sparse Label Correction}
\label{sec:sparse-label-correction}

To address the problem of missing annotations causing false negatives during training, we employ a pseudo-labeling scheme that adds confident predictions to the incomplete set of given targets. 
The inputs for our pseudo-label generation are the teacher predictions $\Phi_T (I_T)$ for an image and the set of available targets, which is in our case the output of the box correction algorithm $C_B$. In the first step, all the predictions with a predicted probability $\hat{s}_p$ of less than a specified threshold $\tau \in [0,1]$ are removed. Next, non-maximum suppression (NMS) is performed on the remaining predicted boxes to remove redundant predictions. For this further reduced set of predicted boxes, the IoUs with the target boxes $c_t$ are computed. All the boxes with a sufficiently large overlap (i.e. $IoU > 0.5$)  with any of the target boxes are removed. Finally, the remaining boxes are added to the new set of target boxes. More formally, the extended and final set of corrected targets $C$ arises as
\begin{align*}
 C= C_B \cup \Big\{
 (\hat{b}_p,\hat{l}_p) |&\exists \hat{s}_p: (\hat{b}_p,\hat{l}_p,\hat{s}_p) \in \textrm{NMS}(\Phi_T(I_T)) \land \hat{s}_p \ge \tau;\\
 &\nexists (c_t,l_t) \in C_B: l_t=\hat{l}_p \land IoU(\hat{b}_p,c_t) > 0.5
 \Big\},
\end{align*}
where $C_B$ is the output of the box correction module (consisting of boxes and labels). This new set of targets $C$ is used as supervision for the student model~$\Phi_S$.

A similar scheme has already been used for sparsely annotated object detection~\cite{co-mining} and for semi-supervised object detection~\cite{unbiased-teacher,soft-teacher,stac}.
An advantage of this method is that it does not modify anything but the targets that are fed into the loss function. The same holds for our box correction method, allowing us to easily combine both in a joint framework.
Analogous to this label mining scheme, a mechanism to drop potential false positive annotations can be implemented. However, we observed that the impact of randomly generated false positive annotations is insignificant compared to the other types of noise discussed in this paper (see Appendix~\ref{sec:app-superfluous}).

\section{Experiments} 
\subsection{Datasets}
\paragraph{Edmonton Trees}
The Edmonton Trees dataset contains urban trees in the city of Edmonton, Canada.
It was created from an inventory\footnote{\url{https://data.edmonton.ca/Environmental-Services/Trees-Map/udbt-eiax}} providing GPS coordinates of trees and an orthomosaic providing RGB imagery\footnote{\url{https://data.edmonton.ca/Thematic-Features/Orthophoto-Repository-2017/3usk-mi6i}} for the area. Tiles of 512 by 512 pixels were extracted from the orthomosaic and matched with the GPS locations of the trees in the inventory. For training, we used 3,536 of such tiles, which contain a total of 25,451 annotated trees. As the annotations were collected on site, trees on private ground are not covered by the inventory and a large number of trees lack annotations. Also, the locations of the annotations are often badly aligned with the trees in the images. Therefore, the Edmonton Trees dataset exhibits sparse and inaccurate annotations. 
Since the images were acquired in a different year than the GPS annotations, the annotations additionally contain many erroneous labels, which further degrades their quality. 
Bounding boxes for training were created from the point annotations by choosing a fixed size of 60 pixels ($\widehat{=}$ 6m) for each side of the box. 
For validation and testing, sets of 128 image tiles were used. To obtain a ground truth, these were labeled manually and directly in the imagery. This resulted in 808 and 863 labeled instances for validation and testing, respectively. In the ground truth annotations, the same fixed size for bounding boxes was used as the detectors are not given the chance to learn to predict boxes of different sizes during training. Furthermore, tree species were not distinguished, i.e. there is only a single category "tree".

With this setup, the average intersection over union of manually labeled ground-truth boxes with their best matching boxes obtained from the inventory amounts to only 0.09. Conversely, the average intersection over union of the boxes from the inventory with their best matching ground-truth boxes amounts to 0.33. The difference in these values can be explained by the missing GPS annotations on private ground. These numbers demonstrate the severity of the noise in the training annotations, making it extremely challenging to robustly train object detectors.
As training on Edmonton Trees was subject to relatively high variance, we report all scores for this dataset as the average and standard deviation over five runs.

\paragraph{NWPU VHR-10}
The NWPU VHR-10~\cite{nwpu} dataset is an optical remote sensing dataset for object detection. It consists of 800 images, of which 150 were selected for validation and testing, respectively. The dataset comprises the ten object classes "bridge", "harbor", "airplane", "ship", "vehicle", "storage tank", "baseball diamond", "tennis court", "basketball court", and "ground track field". As the annotations were created manually and match the images well, we introduce synthetic noise on the training split. More precisely, we follow~\cite{ca-bbc} and displace every horizontal box coordinate by a number of pixels randomly chosen from the interval $[-w N_b, +w N_b]$, where $w$ denotes the pixel width of the bounding boxes and $N_b$ is the box noise level. Analogously, every vertical box coordinate is moved by a random number of pixels in the interval $[-h N_b, +h N_b]$, where $h$ is the height of the box. We report results for the three different box noise levels $N_b = 0\%$, $N_b = 20\%$ and $N_b =40\%$. Furthermore, we sparsify the annotations by randomly removing box annotations. In the settings $N_s=0\%$ and $N_s=50\%$, the corresponding fraction of annotations is dropped. Additionally, we introduce another extreme level of sparse annotations $N_s=ex.$, where only one annotation per image is kept.

\paragraph{Pascal VOC}
Pascal VOC~\cite{pascal-voc} is a benchmark dataset for object detection that is widely used in the computer vision community. We add this dataset to our experiments to be able to directly compare our bounding box correction method with the scores of state-of-the-art methods for bounding box correction, which were reported on Pascal VOC.
Following the setting of \cite{ca-bbc,nar}, we use the union of VOC 2007 \emph{trainval} and VOC 2012 \emph{trainval} for training and test our models on VOC 2007 \emph{test}. From the training images, 10\% were separated and used as a validation set. On the remaining training samples, we manipulate the clean annotations in the same way as for NWPU VHR-10 described above.

\subsection{Implementation Details} 
\label{sec:implementation}
Our code base is built upon Detectron2~\cite{detectron2}. All experiments were conducted on a single NVIDIA Quadro RTX 8000 or a comparable device. Unless specified differently, we employed the Faster R-CNN~\cite{faster-rcnn} architecture with a ResNet-50~\cite{resnet} backbone. Models were trained with SGD with a batch size of 16, a learn rate of 0.02, a momentum of 0.9, and a weight decay of 1e-4. Furthermore, we applied early stopping. On Pascal VOC, the initial learn rate was decreased by a factor of 10 after 20k and 32k iterations in standard training and after 12k and 20k iterations in teacher-student training, whereas we did not see improvements from learn rate decay on the other two datasets.

For teacher-student training, we initialized the detectors with the weights obtained from the best validation iteration for the corresponding setting in standard training. 
Batch sizes were reduced to 8 as two differently augmented versions of every image have to be processed. Apart from that, we kept the same configuration. 
For strong augmentation, we applied brightness and contrast modification, color jitter, and flipping. In contrast to~\cite{unbiased-teacher}, we omitted random erasing and blurring as we did not observe positive effects from that. The EMA keep rate $\alpha$ was chosen to be 0.9996 for Pascal VOC, 0.99 for NWPU VHR-10, and 0.95 for Edmonton Trees (smaller values because of a shorter training schedule).
We did not observe improvements with focal loss (as reported in~\cite{unbiased-teacher}), which is why we left the Faster R-CNN losses unchanged.

For our correction module, a softmax temperature of $\gamma=0.2$ was used in all experiments. The distance limit $d$ and the mining threshold $\tau$ were chosen depending on the noise setting (see Appendix~\ref{sec:app-hyp}). Since the boxes on Edmonton Trees are always of the same size, we modified our box correction method in two ways to account for that. First, instead of using $\delta(b_1, b_2)= 1 - IoU(b_1, b_2)$ as a box distance measure, we used the euclidean distance of the box centers normalized by the standard box size of 60 pixels. On the other hand, the updates of corrected boxes were not performed by averaging all four box coordinates, but by averaging the box center coordinates and expanding the corrected box centers to boxes with the standard size of 60 by 60 pixels. This ensures, that the corrected boxes still conform with the square shape and standard size of target and ground-truth boxes.

\subsection{Main Results}
\label{sec:main-results}
\begin{table}
\centering
\caption{Main results ($AP_{50}$).}
\label{tab:main-results}
\begin{tabular}{ccc@{\,}cc@{\,}c}
\multicolumn{6}{c}{(a) Synthetic annotation noise.} \\
\addlinespace
\toprule
\multicolumn{2}{c}{\textbf{Dataset}} & \multicolumn{2}{c}{\textbf{NWPU VHR-10}} & \multicolumn{2}{c}{\textbf{Pascal VOC}} \\ \midrule
\textbf{$N_b$} & \textbf{$N_s$} & \textbf{Vanilla} & \textbf{Ours} & \textbf{Vanilla} & \textbf{Ours} \\ \midrule
     & 0\%   & 94.2 & 95.8 & 80.8 & --   \\
0\%  & 50\%  & 85.2 & 92.9 & 71.8 & 77.6 \\
     & $ex.$ & 77.3 & 87.9 & 64.5 & 76.3 \\ \midrule
     & 0\%   & 92.3 & 94.8 & 76.6 & 79.8 \\
20\% & 50\%  & 83.2 & 91.9 & 64.9 & 73.8 \\
     & $ex.$ & 53.5 & 77.0 & 59.7 & 73.6 \\ \midrule
     & 0\%   & 72.0 & 92.1 & 58.8 & 77.3 \\
40\% & 50\%  & 50.3 & 67.2 & 45.2 & 66.3 \\
     & $ex.$ & 33.0 & 73.3 & 40.8 & 66.9 \\ \bottomrule
\end{tabular}
\hspace{1cm}
\begin{tabular}{cc}
\multicolumn{2}{c}{(b) Real-world annotation noise.} \\
\addlinespace
\toprule
\multicolumn{2}{c}{\textbf{Edmonton Trees}} \\ \midrule 
\textbf{Vanilla} & \textbf{Ours} \\ \midrule
42.7 $\pm$ 2.2 & 79.8 $\pm$ 1.3 \\
\bottomrule
\end{tabular}
\end{table}

In Table~\ref{tab:main-results}, we provide the results for the different datasets and noise levels in our experiments. When comparing our training method with the standard training of Faster R-CNN, we observe large gains in every setting. Interestingly, the differences between $N_s=50\%$ and $N_s=ex.$ are mostly rather small for our method, while they are substantial for standard training. On Edmonton Trees with real-world annotation noise, the improvement of our method is particularly remarkable, achieving an average of 79.8\% and a maximum of 81.2\% $AP_{50}$ over five runs. These values indicate that our method is indeed capable of ensuring robust training in real-world applications.

Furthermore, we applied our method in the clean setting ($N_b=0\%$, $N_s=0\%$). For NWPU VHR-10, we observe a small improvement of 1.6 points in $AP_{50}$. However, when we removed the correction module in the teacher-student training, we observed an $AP_{50}$ gain of 1.8 points. For Pascal VOC, the performance dropped after initialization with the vanilla model -- independent of whether the correction module was used or omitted. Thus, we cannot conclude that generating corrected pseudo-labels with our correction module leads to improvements when clean annotations are available.

\subsection{Ablation Study} 

\begin{table*}[t]
\centering
\caption{Ablation study ($AP_{50}$).}
\label{tab:ablation}
\begin{tabular}{ccc@{\hspace{15mm}}ccc}
\toprule
\begin{tabular}[c]{@{}c@{}}\textbf{Teacher-} \\ \textbf{student}\end{tabular} & \begin{tabular}[c]{@{}c@{}}\textbf{Box} \\ \textbf{Correction}\end{tabular} & \begin{tabular}[c]{@{}c@{}}\textbf{Sparse Label}\\ \textbf{Correction}\end{tabular} & \begin{tabular}[c]{@{}c@{}}\textbf{Edmonton} \\ \textbf{Trees}\end{tabular} & \begin{tabular}[c]{@{}c@{}}\textbf{NWPU VHR-10} \\ $(N_b=40\%, N_s=ex.)$ \end{tabular} & \begin{tabular}[c]{@{}c@{}}\textbf{Pascal VOC} \\ $(N_b=40\%, N_s=ex.)$ \end{tabular}  \\ \midrule
\xmark & \xmark & \xmark & 42.7 $\pm$ 2.2  & 33.0 & 40.8\\
\cmark & \cmark & \xmark & 57.5 $\pm$ 2.2  & 72.1 & 65.4       \\
\cmark & \xmark & \cmark & 63.1 $\pm$ 11.0 & 46.6 & 51.9       \\
\xmark & \cmark & \cmark & NaN             & 38.5 & -- \\
\cmark & \cmark & \cmark & \textbf{79.8 $\pm$ 1.3}  & \textbf{73.3} & \textbf{66.9}       \\ \bottomrule
\end{tabular}
\end{table*}
To further show the effects of individual parts of our method, we present an ablation study in Table~\ref{tab:ablation}. In every setting, the standard Faster R-CNN architecture is employed. The first row of the table corresponds to the vanilla models in Table~\ref{tab:main-results}. The second row provides scores when training with box correction but without the sparse label correction module that adds pseudo-labels to the targets. Conversely, the third row only employs this sparse label correction module and the box correction is omitted. Both mechanisms alone can achieve substantial performance gains. Remarkably, the magnitudes of the gains are different for the datasets. We suppose that missing annotations have more severe effects than imprecise localizations on Edmonton Trees. However, using the teacher-student framework and both mechanisms in combination (last row) clearly leads to the best results. When we do not perform teacher-student training, but the student model is supervised by its own target corrections (third row), the performance on Pascal VOC worsens right after switching on the correction mechanism. For Edmonton Trees, training even diverges. Hence, teacher-student training is critical for stable training in our approach. An error analysis for these ablations can be found in Appendix~\ref{sec:app-tide}.

\subsection{State-of-the-art Comparison for Box Correction on Pascal VOC}
Since there are no related works dealing with severe bounding box noise and sparse annotations simultaneously, we cannot conduct a direct state-of-the-art comparison. However, there are some publications reporting scores for simulated bounding box noise on Pascal VOC. Thus, we compare our box correction module in combination with the presented teacher-student training with their methods. As we can see in Table~\ref{tab:voc-sota}, our model does not only surpass the vanilla Faster R-CNN, but also all other approaches. At the same time, our method does not require modifying the detector architecture, making it easier to apply and extend.

\subsection{Detector Architecture Comparison} 

As our method can be applied with different detector architectures, we investigate the effect of the architecture in Table~\ref{tab:detector-comparison}. We conduct experiments with the two-stage detector Faster R-CNN~\cite{faster-rcnn}, the one-stage detector RetinaNet~\cite{retinanet}, and the anchor-free one-stage detector FCOS~\cite{fcos}.

Apparently, all three detectors perform similarly if clean annotations and vanilla training are used. For standard training and noisy annotations, RetinaNet seems to outperform the other architectures. Interestingly, this does not hold if we apply our noise-robust training scheme. Here, Faster R-CNN and RetinaNet are basically on par, while FCOS is slightly outperformed. Nonetheless, all methods greatly benefit from our training method, showing that is it in principle applicable for various detector architectures.

\begin{table}
\centering
\caption{SOTA comparison on PascalVOC ($AP_{50}$). Results in the second table section were taken from \cite{ca-bbc,nar}. Our scores were averaged from three runs with different seeds for noise simulation.}
\label{tab:voc-sota}
\begin{tabular}{lc@{\quad}c}
\toprule
\multirow{2}{*}{\textbf{Method}} & \multicolumn{2}{c}{\textbf{Box Noise Level $N_b$}} \\  & 20\%  & 40\% \\ \midrule
Vanilla Faster R-CNN \cite{faster-rcnn} & 76.6 $\pm$ 0.3          & 58.8 $\pm$ 0.8 \\ \midrule
Co-Teaching \cite{co-teaching,ca-bbc}   & 75.6          & 60.6 \\
SD-LocNet \cite{sd-locnet,ca-bbc}       & 75.3          & 59.7 \\
NOTE-RCNN \cite{note-rcnn,ca-bbc}       & 76.0          & 63.4 \\
CA-BBC \cite{ca-bbc}                    & 77.9          & 71.9 \\
NAR \cite{nar}                          & 78.4          & 73.4 \\ \midrule
Ours (Faster R-CNN)                     & \textbf{79.8 $\pm$ 0.4} & \textbf{77.3 $\pm$ 0.2} \\ \bottomrule
\end{tabular}
\end{table}

\begin{table}
\centering
\caption{Comparison of different detector architectures when training on Pascal VOC with $N_b=40\%$ and $N_s=ex.$ ($AP_{50}$)}
\label{tab:detector-comparison}
\begin{tabular}{cccc}
\toprule
\textbf{Architecture}             & \begin{tabular}[c]{@{}c@{}}\textbf{Vanilla}\\ (noisy)\end{tabular} & \begin{tabular}[c]{@{}c@{}}\textbf{Ours}\\ (noisy)\end{tabular} & \begin{tabular}[c]{@{}c@{}}\textbf{Vanilla}\\ (clean)\end{tabular} \\ \midrule
Faster R-CNN & 40.8          & \textbf{66.9} & \textbf{80.8} \\
RetinaNet    & \textbf{45.4} & 66.4          & 80.0          \\
FCOS         & 40.4          & 64.7          & 80.2          \\ \bottomrule
\end{tabular}
\end{table}

\subsection{Qualitative Results} 

\begin{figure*}[t!]
    \centering
    \begin{subfigure}{0.85\textwidth}
        \centering    
        \begin{subfigure}{0.32\columnwidth}
            \centering
             \includegraphics[width=\columnwidth]{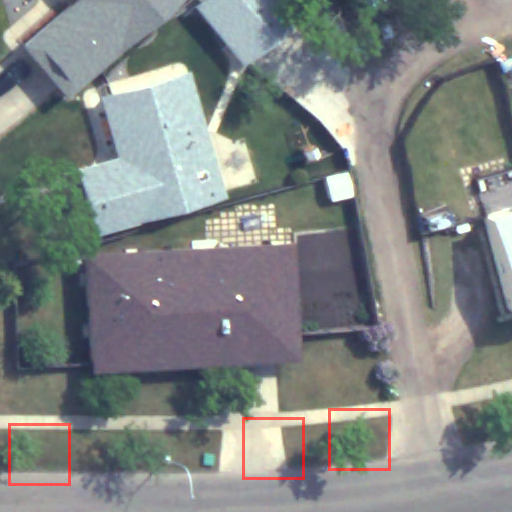}
        \end{subfigure}
        \begin{subfigure}{0.32\columnwidth}
            \centering
             \includegraphics[width=\columnwidth]{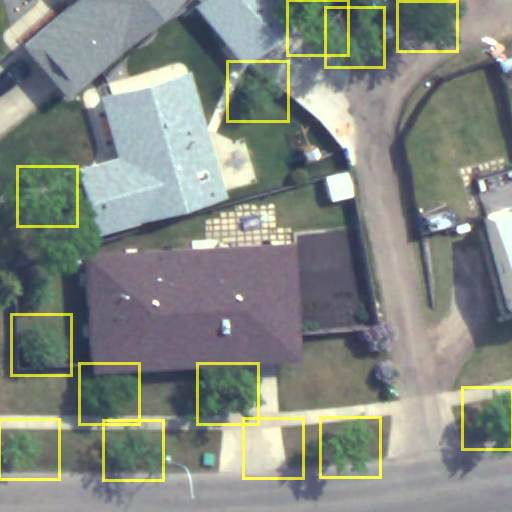}
        \end{subfigure}
        \begin{subfigure}{0.32\columnwidth}
            \centering
             \includegraphics[width=\columnwidth]{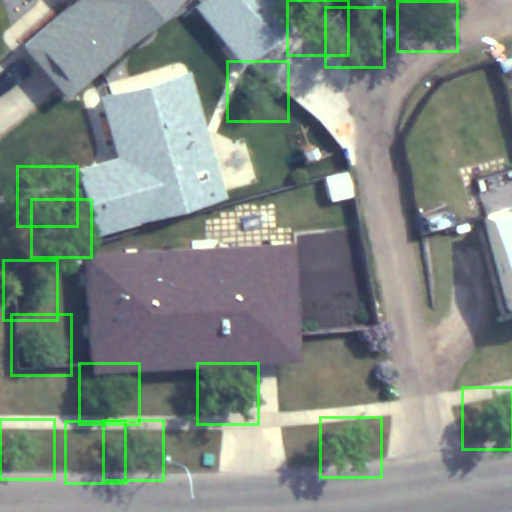}
        \end{subfigure}
        \subcaption{Edmonton Trees}
    \end{subfigure}    
    
    \begin{subfigure}{0.85\textwidth}
        \centering
        \begin{subfigure}{0.32\columnwidth}
            \centering
             \includegraphics[width=\columnwidth]{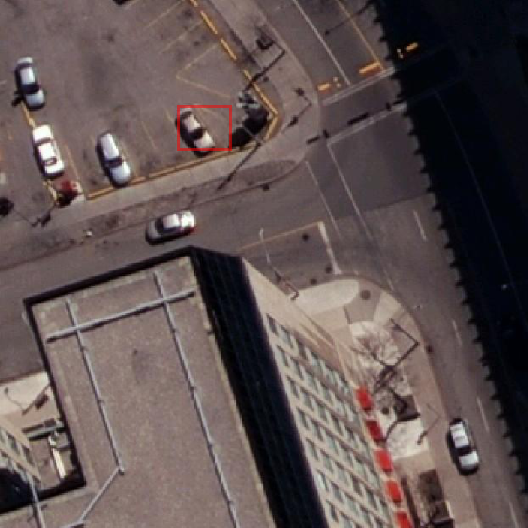}
        \end{subfigure}
        \begin{subfigure}{0.32\columnwidth}
            \centering
             \includegraphics[width=\columnwidth]{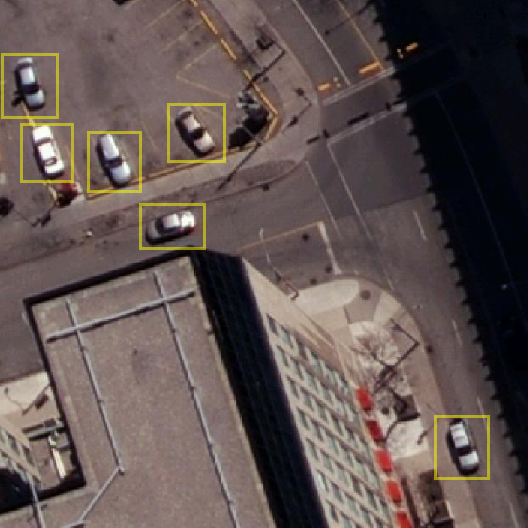}
        \end{subfigure}
        \begin{subfigure}{0.32\columnwidth}
            \centering
             \includegraphics[width=\columnwidth]{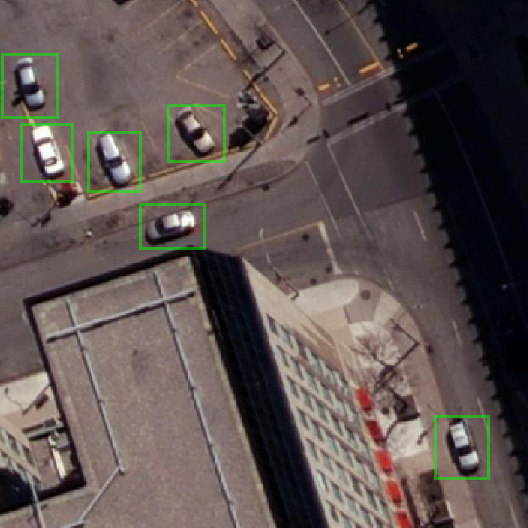}
        \end{subfigure}
        \subcaption{ NWPU VHR-10} 
    \end{subfigure}
    
    \caption{Qualitative examples. Noisy annotations (left), corrected annotations (middle) and final predictions(right) \emph{Best viewed in color.}}
    \label{fig:examples}
\end{figure*}

In Figure~\ref{fig:examples}, we provide some visual examples for our method. As we can see, our correction method manages to refine the noisy annotations to a satisfactory extent. Accordingly, the final predictions of the model are of substantially higher quality than the initial annotations. Figure~\ref{fig:examples} (a) additionally demonstrates the poor annotation quality in the Edmonton Trees dataset. Due to our training method, the model still accomplishes to produce good results.

Nonetheless, we observed a recurring error type with our method. If a bounding box is extremely badly placed such that the overlap with its object becomes very small, the box correction mechanism may not manage to correct its location properly. Instead, we observed cases where a new box that matches the object better is added to the set of targets, leading to two target boxes representing one object. We tried to avoid this behavior by choosing higher values for the distance limit $d$ or by introducing a mechanism to drop erroneous boxes, but we did not observe gains. We conclude that there is a trade-off between the flexibility of the correction module and training stability. 

\section{Conclusion} 
In this paper, we proposed a method for training object detectors in (extremely) noisy settings with incomplete and imprecise bounding box annotations. Its simple design and its effectiveness on both simulated and real-world annotation noise make it valuable in many practical scenarios where clean annotations are not available.
We are convinced that the remote sensing community benefits from our work, as our method allows robust training of object detectors without the necessity of manually labeling large amounts of images, therefore removing a considerable barrier for employing object detection in practice.
We did not address the effects of class label noise in this work. However, we argue that the design of our method allows adding appropriate correction mechanisms to account for that as well. Furthermore, a requirement for our method is that the noisy annotations alone suffice to learn the features of interest to a certain extent. We see it as our future work to combine large amounts of noisy annotations with small amounts of clean annotations in order to cope with situations where this requirement becomes a limitation. Also, the combination with a subset of clean annotations could further close the gap to settings with complete and clean annotations.

\section*{Acknowledgements}
This work has been funded by the German Federal Ministry of Education and Research (BMBF) under Grant No. 01IS18036A. The authors of this work take full responsibilities for its content. Additionally, we thank Guillermo Castilla for the fruitful discussion in the early phase of this project.

\bibliographystyle{plain} 

\appendix 

\section{Error Analysis with TIDE}
\label{sec:app-tide}
\begin{figure}[b]
    \centering
    \includegraphics[width=0.24\columnwidth]{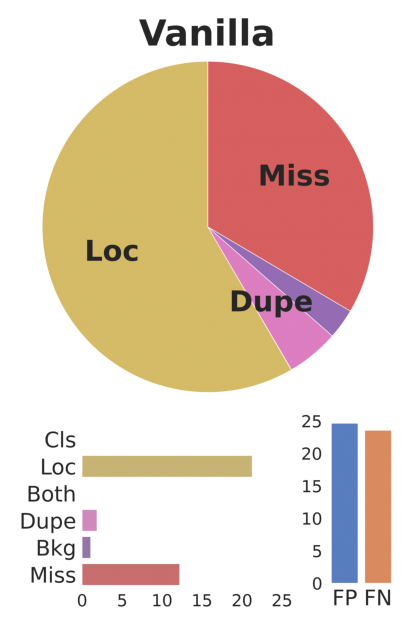}
    \includegraphics[width=0.24\columnwidth]{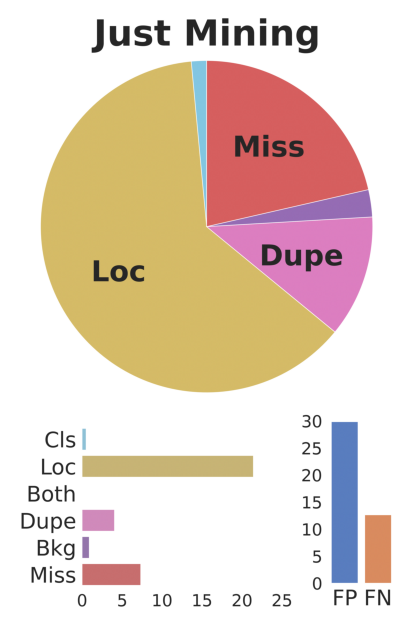}
    \includegraphics[width=0.24\columnwidth]{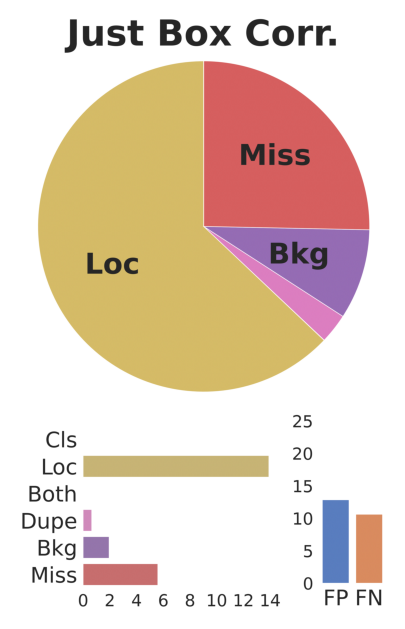}
    \includegraphics[width=0.24\columnwidth]{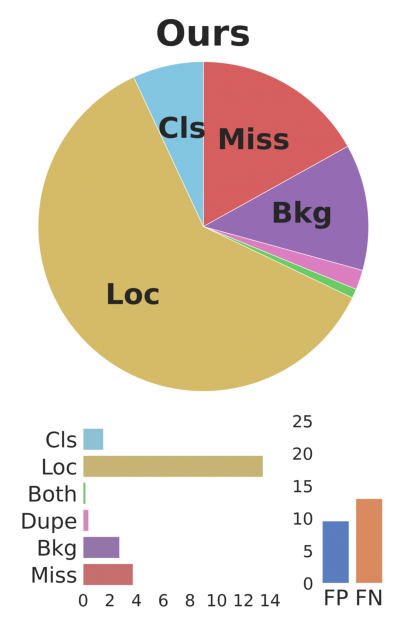}
    \caption{TIDE~\cite{tide} plots for vanilla Faster R-CNN and our method with different ablations, obtained on NWPU VHR-10 ($N_b=40\%, N_s=ex.$). \emph{Best viewed in color.}}
    \label{fig:tide}
\end{figure}

TIDE~\cite{tide} is a toolbox for analyzing error sources in object detection. We show visualizations created with TIDE in Figure~\ref{fig:tide}. The plot in the top left corresponds to Faster R-CNN~\cite{faster-rcnn} with standard training and the plot in the bottom right corresponds to our full method including teacher-student training and target correction. Analogous to the ablation study in Table~\ref{tab:ablation}, the plots in the top right and the bottom left correspond to our method where only the label mining and the box correction were applied, respectively.

From this, we can gain the following insights:
If we add our mining mechanism (top right), the number of missing predictions and false negatives is significantly reduced compared to the vanilla Faster R-CNN (top left). If we only add our box correction mechanism (bottom left), the number of localization errors is largely reduced, while we interestingly also observe an improvement in missing prediction errors compared to the vanilla model. Comparing the model with box correction only and our full model (bottom right), we observe a similar overall performance (72.1\% vs. 73.3 \% $AP_{50}$, see Table~\ref{tab:ablation}). The advantage of the full method can be explained by the reduction of missing prediction errors, which is caused by the mining module. However, when applying the mining module, we can see a slight tendency to produce classification errors. This might occur because incorrect pseudo-labels are mined and used for supervision. Nonetheless, the effect of these errors appear to be insignificant.

\begin{figure}
    \centering
    \includegraphics[width=0.9\columnwidth]{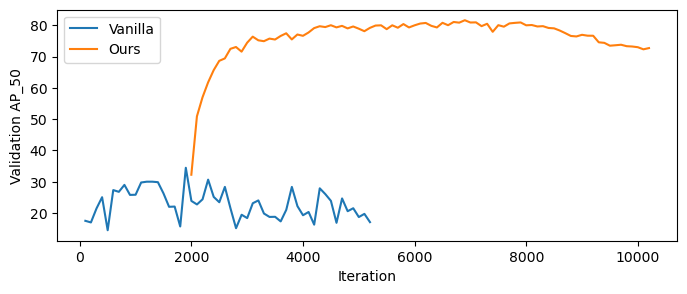}
    \caption{Validation $AP_{50}$ curves on Edmonton Trees for standard training and our training method. \emph{Best viewed in color.}}
    \label{fig:curves}
\end{figure}

\section{Training on Edmonton Trees} 

\begin{table*}
\centering
\caption{Correction hyperparameters for the main results (Table \ref{tab:main-results})}
\label{tab:hyp-main-results}
\begin{tabular}{cccc@{\quad}ccc@{\quad}cccc}
\toprule
\textbf{Dataset}     & \textbf{\begin{tabular}[c]{@{}c@{}}Edmonton \\ Trees\end{tabular}} & \multicolumn{9}{c}{\textbf{NWPU VHR-10 \& Pascal VOC}}                                       \\ \midrule
Box Noise Level $N_b$   & ?     & \multicolumn{3}{c}{0\%} & \multicolumn{3}{c}{20\%} & \multicolumn{3}{c}{40\%}\\
Sparsity Level $N_s$    & ? & 0\%    & 50\%   & $ex.$ & 0\%    & 50\%    & $ex.$ & 0\%    & 50\%    & $ex.$ \\ \midrule
$d$                     & 0.5 & 0.1  & --   & --   & 0.35 & 0.35 & 0.35  & 0.6   & 0.6   & 0.6 \\ 
$\tau$                  & 0.8 & 0.95 & 0.9  & 0.8  & --   & 0.9  & 0.8   & --    & 0.8   & 0.8 \\ \bottomrule
\end{tabular}
\end{table*}

\begin{table}
\centering
\caption{Hyperparameters for the detector comparison (Table~\ref{tab:detector-comparison})}
\label{tab:hyp-detector-comparison}
\begin{tabular}{cc@{\quad}c@{\quad}c}
\toprule
\textbf{Detector}   & lr    & $d$  & $\tau$ \\ \midrule
Faster R-CNN        & 0.02  & 0.6  & 0.8    \\
RetinaNet           & 0.01  & 0.6  & 0.4    \\
FCOS                & 0.01  & 0.6  & 0.5    \\ \bottomrule
\end{tabular}
\end{table}
As we mentioned in the paper, training on Edmonton Trees is rather unstable because of the strong annotation noise. This holds especially for standard training. As we can see in Figure~\ref{fig:curves}, the vanilla model gets stuck in a regime with high variance and low detection scores. However, when we train with our method, the model -- initialized from the best iteration in vanilla training -- immediately gets out of this regime and greatly improves in $AP_{50}$. On top of that, the variance across iterations is largely reduced due to EMA updating. In the late training phase, we observe overfitting with our method.

\section{Qualitative Comparison of Standard Training and Our Method} 

\label{sec:app-qual}
\begin{figure}[t!]
    \centering
    \begin{subfigure}{\columnwidth}
        \centering
        \includegraphics[width=0.4\columnwidth]{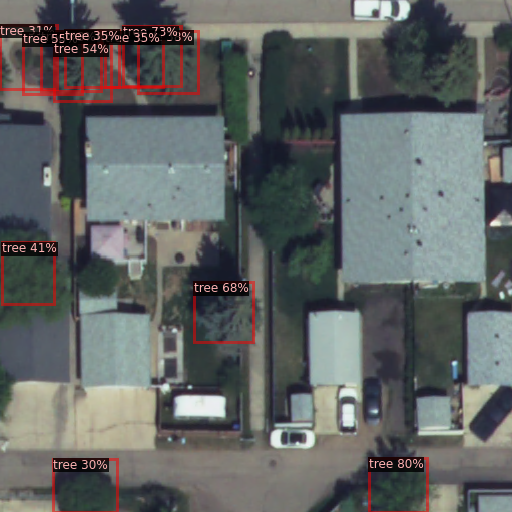}
        \includegraphics[width=0.4\columnwidth]{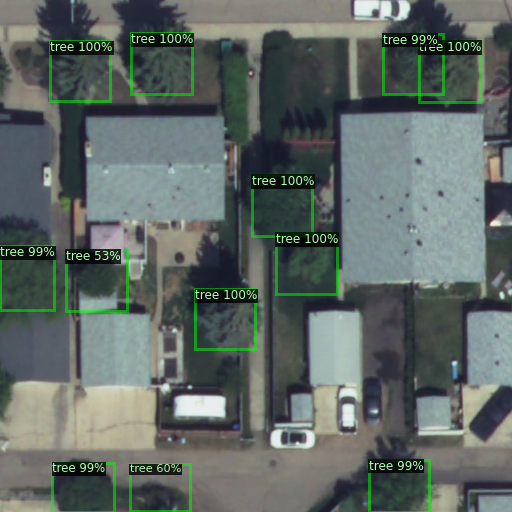}
        \subcaption{Edmonton Trees}
    \end{subfigure}

    \begin{subfigure}{\columnwidth}
        \centering
        \includegraphics[width=0.4\columnwidth]{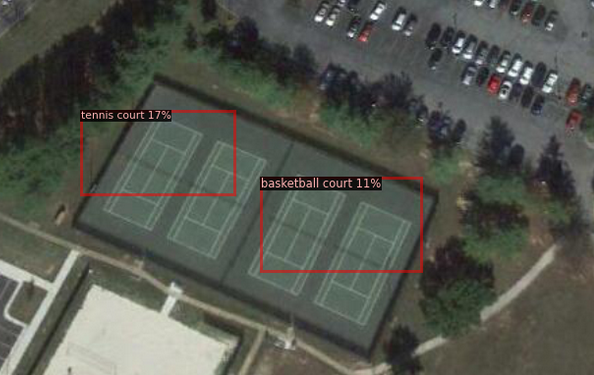}
        \includegraphics[width=0.4\columnwidth]{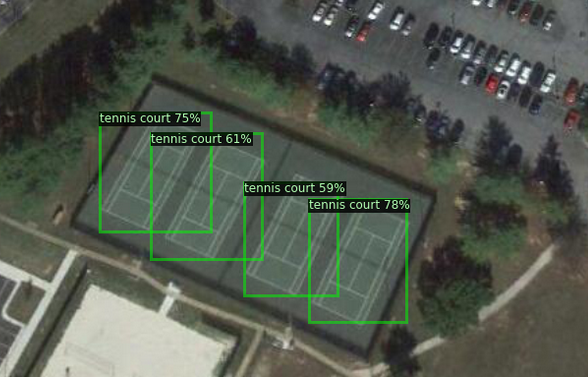}
        \subcaption{NWPU VHR-10}
    \end{subfigure}

    \begin{subfigure}{\columnwidth}
        \centering
        \includegraphics[width=0.4\columnwidth]{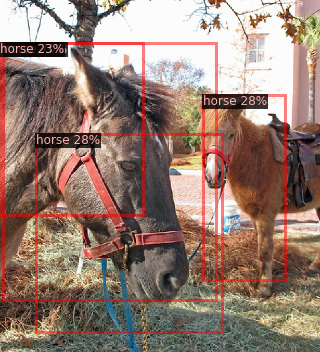}
        \includegraphics[width=0.4\columnwidth]{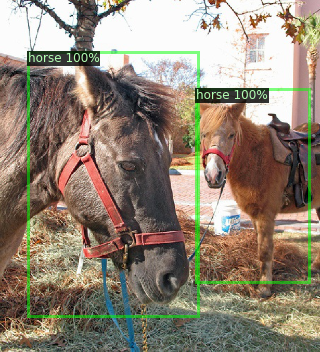}
        \subcaption{Pascal VOC ($N_b=40\%$, $N_s=ex.$)}
    \end{subfigure}
    \caption{Qualitative examples for Faster R-CNN with standard training (left) and training with our method (right). \emph{Best viewed on screen.}}
    \label{fig:supp-examples}
\end{figure}
We show a visual comparison of our method and standard training of Faster R-CNN in Figure~\ref{fig:supp-examples}. On the left-hand side, we can see that, with standard training, the model fails to accurately place the bounding boxes. This is caused by the localization noise in the supervision. Also, the confidence scores are relatively low and some instances are not detected at all. In contrast, the predictions obtained with our method (right-hand side) demonstrate a much better ability to localize objects. The main reason for that is our correction mechanism allowing the model to learn to distinguish between properly and poorly placed boxes.

\section{Effect of False Positive Annotations} 

\label{sec:app-superfluous}
To quantify the effect of superfluous boxes in the annotations, we artificially added boxes to the NWPU VHR-10 dataset. For that, we sampled the number of superfluous boxes for every image from a binomial distribution with $n=10$ and $p=0.5$. The locations and class labels were chosen randomly and uniformly. Also, the box heights and widths were sampled independently and uniformly with a minimum and maximum size of 16 and 196 pixels, respectively. Overall, this resulted in 5,117 boxes in the training split, compared to 2,557 boxes in the original, clean dataset. When training a Faster R-CNN model with standard training, we observed an $AP_{50}$ of $94.0 \%$. Hence, the performance dropped by only 0.2 points in comparison with the same model trained on the clean dataset ($94.2 \% AP_{50}$). We conclude that the effect of false positive supervision has marginal effects on standard object detectors, which is why we do not further investigate in this direction. 

\section{Hyperparameters} 

\label{sec:app-hyp}

In Table~\ref{tab:hyp-main-results}, we provide the correction hyperparameters, namely the distance limit $d$ and the mining threshold $\tau$, used to obtain our main results. With "--", we indicate that the respective submodule was not used in this setting. The rest of the hyperparameters and configurations are specified in the implementation details (Section~\ref{sec:implementation}).

Furthermore, we provide the hyperparameters for the detector comparison in Table~\ref{tab:hyp-detector-comparison}. The parameters not listed in Table~\ref{tab:hyp-detector-comparison} were chosen identical to the ones described in the implementation details for all three detector architectures. Here, it is notable that the one-stage detectors RetinaNet~\cite{retinanet} and FCOS~\cite{fcos} had far lower confidence scores in their predictions. Therefore, we needed to lower the mining threshold $\tau$ such that a reasonable number of pseudo-boxes was mined.

\end{document}